\documentclass{article}

\usepackage{microtype}
\usepackage{graphicx}
\usepackage{subfigure}
\usepackage{booktabs} % for professional tables

\usepackage{hyperref}

\usepackage[accepted]{icml2024}
\usepackage{amsmath}
\usepackage{amssymb}
\usepackage{mathtools}
\usepackage{amsthm}
\usepackage{epigraph}
\usepackage{etoolbox}
\usepackage{tabularx}
\usepackage{enumitem}
\setlist[itemize]{align=parleft,left=0pt..1em}
\setlist[enumerate]{align=parleft,left=0pt..1em}
\usepackage[font={small}]{caption}

\usepackage[capitalize,noabbrev]{cleveref}

\theoremstyle{plain}

\theoremstyle{definition}

\theoremstyle{remark}

\usepackage[textsize=tiny]{todonotes}
\definecolor{chestnut}{cmyk}{0, 0.7808, 0.4429, 0.1412}

\icmltitlerunning{On the Limitations and Prospects of Machine Unlearning for Generative AI}

\begin{document}

\twocolumn[
\icmltitle{On the Limitations and Prospects of Machine Unlearning for Generative AI}

% It is OKAY to include author information, even for blind
% submissions: the style file will automatically remove it for you
% unless you've provided the [accepted] option to the icml2024
% package.

% List of affiliations: The first argument should be a (short)
% identifier you will use later to specify author affiliations
% Academic affiliations should list Department, University, City, Region, Country
% Industry affiliations should list Company, City, Region, Country

% You can specify symbols, otherwise they are numbered in order.
% Ideally, you should not use this facility. Affiliations will be numbered
% in order of appearance and this is the preferred way.
\icmlsetsymbol{equal}{*}

\begin{icmlauthorlist}
\icmlauthor{Shiji Zhou}{thu}
\icmlauthor{Lianzhe Wang}{thu}
\icmlauthor{Jiangnan Ye}{ic}
\icmlauthor{Yongliang Wu}{seu}
\icmlauthor{Heng Chang}{thu}
%\icmlauthor{}{sch}
%\icmlauthor{}{sch}
%\icmlauthor{}{sch}
\end{icmlauthorlist}

\icmlaffiliation{thu}{Tsinghua University}
\icmlaffiliation{seu}{Southeast University}
\icmlaffiliation{ic}{Imperial College London}

\icmlcorrespondingauthor{Heng Chang}{changh17@tsinghua.org.cn}

% You may provide any keywords that you
% find helpful for describing your paper; these are used to populate
% the "keywords" metadata in the PDF but will not be shown in the document
\icmlkeywords{Machine UnLearning, Generative AI}

\vskip 0.3in
]

% this must go after the closing bracket ] following \twocolumn[ ...

% This command actually creates the footnote in the first column
% listing the affiliations and the copyright notice.
% The command takes one argument, which is text to display at the start of the footnote.
% The \icmlEqualContribution command is standard text for equal contribution.
% Remove it (just {}) if you do not need this facility.

% \printAffiliationsAndNotice{}  % leave blank if no need to mention equal contribution
\printAffiliationsAndNotice{\icmlEqualContribution} % otherwise use the standard text.

\begin{abstract}
Generative AI (GenAI), which aims to synthesize realistic and diverse data samples from latent variables or other data modalities, has achieved remarkable results in various domains, such as natural language, images, audio, and graphs. However, they also pose challenges and risks to data privacy, security, and ethics. Machine unlearning is the process of removing or weakening the influence of specific data samples or features from a trained model, without affecting its performance on other data or tasks. While machine unlearning has shown significant efficacy in traditional machine learning tasks, it is still unclear if it could help GenAI become safer and aligned with human desire. To this end, this position paper provides an in-depth discussion of the machine unlearning approaches for GenAI. Firstly, we formulate the problem of machine unlearning tasks on GenAI and introduce the background. Subsequently, we systematically examine the limitations of machine unlearning on GenAI models by focusing on the two representative branches: LLMs and image generative (diffusion) models. Finally, we provide our prospects mainly from three aspects: benchmark, evaluation metrics, and utility-unlearning trade-off, and conscientiously advocate for the future development of this field.
\end{abstract}
\section{Introduction}

\epigraph{\textit{“Remembrance is a form of meeting. Forgetfulness is a form of freedom.”}}{Kahlil Gibran (1926)}

% \lianzhe{below is a semi-auto generated intro, version 1}

In an era marked by the burgeoning influence of Generative AI  (GenAI) \cite{baidoo2023education}, we are rapidly progressing toward a digital future dominated by AI-generated content. This technological advancement has become a cornerstone in various domains, including natural language processing, image synthesis, audio generation, and graph-based applications. While GenAI heralds an era of innovation and efficiency, it simultaneously raises pressing concerns about data privacy, security, and ethical implications \cite{carlini2023extracting}.

The training datasets employed in GenAI often contain sensitive information encompassing private, copyrighted, or potentially harmful content \cite{dubinski2024towards}. This situation raises significant risks of sensitive data leakage \cite{wu2022membership}, directly conflicting with the growing legislative emphasis on the ``right to be forgotten'' \cite{rosen2011right}. Instances such as the proliferation of copyright infringement cases post the release of models like Stable Diffusion \cite{rombach2022high}, and The New York Times's lawsuit against OpenAI for content leakage\footnote{https://nytco-assets.nytimes.com/2023/12/NYT$\_$Complaint$\_$Dec2023.pdf}, underscore the urgency of addressing these issues.

In response to these challenges, Machine unlearning \cite{bourtoule2021machine} has emerged as a potentially promising solution. Machine unlearning aims to compel models to forget sensitive information, thereby fundamentally eliminating the risk of content leakage. This approach, which seeks to erase sensitive memories directly, stands in contrast to filtering-based solutions that are often susceptible to bypassing or direct attacks. Current research on Machine unlearning spans various generative models, including Large Language Models (LLMs) \cite{yao2023large}, image generative models \cite{mishkin2022dall}, and multi-modal generative models \cite{suzuki2022survey}. These studies have demonstrated machine unlearning's potential in removing elements like copyrighted styles \cite{gandikota2023erasing}, fictional characters \cite{eldan2023s}, and private data \cite{tarun2023fast}.

\begin{figure}[t]
  \centering
  \includegraphics[width=\columnwidth]{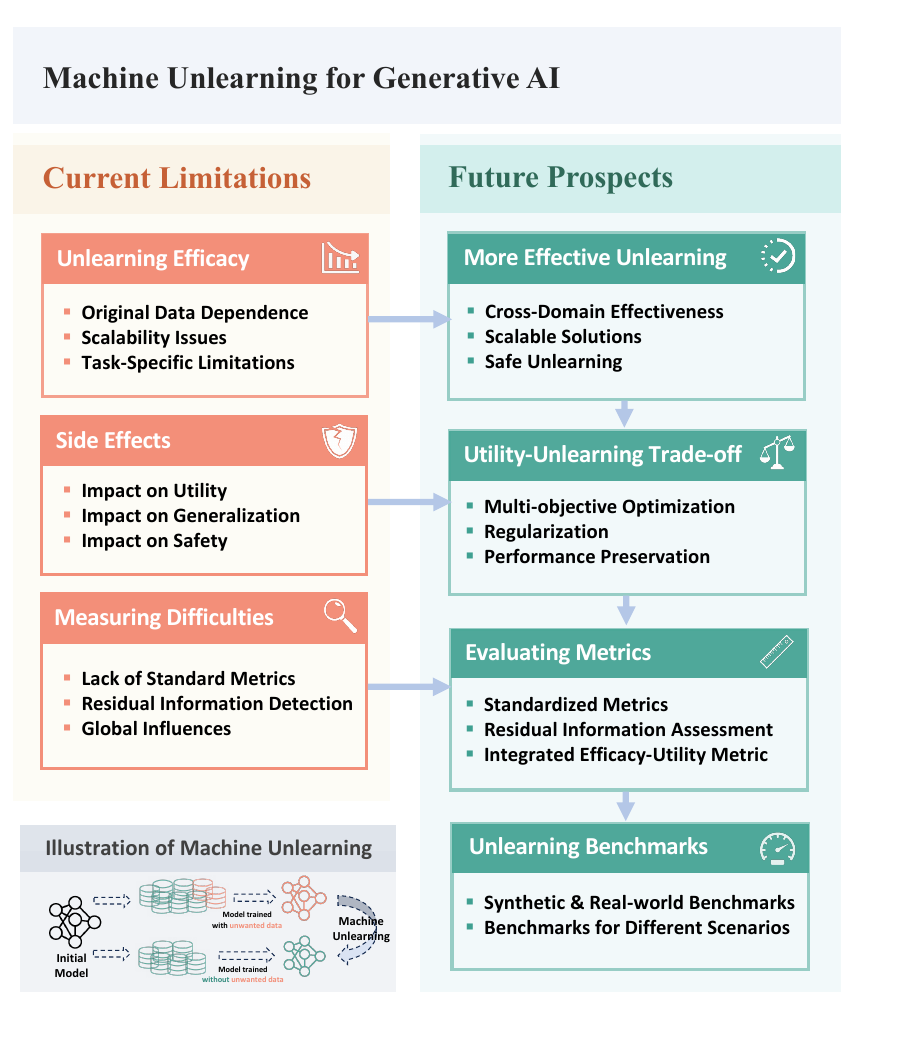}
  \vspace{-8mm}
  \caption{Summarization of our position on the limitations and prospects of machine unlearning methods on GenAI.}
  \label{fig:motiv}
  \vspace{-6mm}
\end{figure}

However, the exploration of MU in the context of GenAI is still nascent, with several limitations hindering its full potential. As illustrated in Figure~\ref{fig:motiv}, we summarise the urgent limitations of the current machine unlearning methods on GAI into three perspectives. Firstly, as an emerging technique, machine unlearning for GenAI is still distant from achieving a level of efficacy requisite for practical applications. Secondly, the evaluation metrics currently employed in MU research are insufficiently robust and fail to capture the multifaceted impact of unlearning on generative models. Thirdly, the side effects of unlearning, including its impact on model performance, generalization, and safety, are significant concerns that must be addressed. As a basis of this position paper, we aim to raise public awareness of these limitations and provide insights that guide future research to address these problems.

In this position paper, we systematically discuss the limitations from the angles of efficacy, side effects, and measurement. Our discussion ranges from LLMs to image generative models. Based on these current inadequacies of unlearning approaches, we advocate future research to focus on three fundamental paths: the benchmarking of unlearning methods, the development of robust and complete evaluating metrics, and the investigation of the balance between utility and unlearning. We believe exploring these three paths should pave the way for further enhancing machine unlearning on GenAI. Overall, our contributions can be summarized as follows:
\begin{itemize}
\item This is, to the best of our knowledge, the first attempt to comprehensively review the limitations of current research on machine unlearning in GenAI. 
\item We offer a systematic examination of the limitations in current unlearning approaches, covering a broad spectrum of models from LLMs to image generative models. This examination provides researchers with a clearer insight into the urgent problems in this area.
\item Tackling these limitations, we propose three prospects based on the examinations, aiming to chart a course for future research in machine unlearning for GenAI.
\end{itemize}

Through this paper, we seek to contribute to the growing body of knowledge in GenAI, specifically in the area of machine unlearning, and catalyze further research that addresses the critical challenges identified.

\section{Background}
\subsection{Unlearning Formulation} 
Let $\mathcal{D}_{tr}=\{(x_i)\}_{i=1}^N$ be the training data where $x_i\in \mathcal{X}$ is the training input. Suppose we have the original model by optimizing towards the training data:
\begin{equation}
    \theta_0 = \arg\min_{\theta\in \mathcal{K}} \mathbb{E}_{x\sim \mathcal{D}_{tr}} \mathcal{L}(\theta,x).
\end{equation}
Let $\mathcal{D}_{f}\subseteq\mathcal{D}_{tr}$ be a subset of training data that is harmful to the model and needs to be forgotten, and $\mathcal{D}_{r} = \mathcal{D}_{tr} \setminus \mathcal{D}_{f}$ be remaining training data of which information we want to retain. The goal of machine unlearning is to successfully unlearn $\mathcal{D}_{f}$, and the \emph{exact unlearning} is to obtain the \emph{gold standard model} retrained from scratch with only $\mathcal{D}_{r}$:
\begin{equation}
    \theta^* = \arg\min_{\theta\in \mathcal{K}} \mathbb{E}_{x\sim \mathcal{D}_{r}} \mathcal{L}(\theta,x).
\end{equation}
Exact unlearning can be obtained by retraining, which causes tremendous cost that is not affordable to frequently updating models. In practice, \emph{approximate unlearning} aims to finetune the original model to obtain the unlearned model $\theta^{u}$ whose output distribution $P_{\theta^{u}}(\cdot)$ approximates the distribution of gold standard model $P_{\theta^*}(\cdot)$. Unless otherwise indicated, we only discuss the approximate unlearning in the next context in this paper.

% Machine unlearning is a relatively new and emerging field, especially for GenAI. In the following we provide an overview of the state-of-the-art (SOTA) machine unlearning methods for two categories of generative models: (1) unlearning for large language models (LLMs), and (2) unlearning for image generative models, respectively.
\subsection{Unlearning for LLM}
Machine unlearning for LLMs is a crucial technique to align LLMs with human preferences and values and to ensure their ethical and responsible use. The existing methods for machine unlearning for LLMs can be broadly classified into:

\paragraph{Parameter Optimization Methods.} These methods update the model parameters by minimizing a loss function that penalizes the undesirable outputs or behaviors of the model. ~\cite{yao2023large} proposed a gradient-based unlearning method that minimizes the cross-entropy loss between the model outputs and a predefined target distribution for the data samples that need to be unlearned. They applied their method to three scenarios of unlearning for LLMs: removing harmful responses, erasing copyright-protected content, and eliminating hallucinations.

\paragraph{Parameter Merging Methods.} These methods reduce the model size and complexity by merging or pruning the model parameters that are most affected by the data samples that need to be unlearned. \cite{ilharco2022editing} proposed the concept of a task vector, which, through arithmetic operations like negation or addition between task vectors, can selectively modify the model’s output with minimal impact on other model behaviors.

\paragraph{In-context Learning Methods.} These methods modify the model inputs or outputs by adding or removing certain tokens or features that indicate the data samples or modalities that need to be unlearned.  To unlearn a particular instance in the forget set, \cite{pawelczyk2023context} provided the instance alongside a flipped label and additional correctly labeled instances which are prepended as inputs to the LLM at inference time. These contexts are shown to be able to effectively remove specific information in given instances while maintaining comparable performance with other unlearning methods that need to access the LLM parameters.

% The main challenges for machine unlearning for LLMs are:
% \begin{itemize}
%     \item The large size and complexity of LLMs make retraining or fine-tuning them from scratch computationally expensive and impractical.
%     \item The lack of explicit labels or feedback for the data samples or features that need to be unlearned makes it difficult to identify and quantify the undesirable effects of the data on the model.
%     \item  The trade-off between unlearning performance and model utility, which requires balancing the degree of forgetting the unwanted data and preserving the generalization ability of the model on other data and tasks.
% \end{itemize}

\subsection{Unlearning for Image Generative Model}
Image generative models have various applications, such as image editing, style transfer, super-resolution, and data augmentation~\cite{iqbal2022survey}. However, image generative models also face challenges and risks, such as violating data privacy, infringing data ownership, and generating inappropriate or misleading images.
Based on the degree of influence removal achieved, the existing methods for machine unlearning for image generative models can be broadly classified into two categories~\cite{xu2023machine}:

\paragraph{Exact Unlearning Methods.} These methods focus on removing the influence of targeted data points from the model through retraining at the algorithmic level completely. It usually involves censoring images from the training dataset such as removing all people's images~\cite{nichol2021glide} or excluding undesirable classes of data and then performing model retraining~\cite{mishkin2022dall}. The retraining process of large models is often costly, which makes this dataset removal-based approach less practical.

\paragraph{Approximate Unlearning Methods.} These methods aim to minimize the influence of target data points in an efficient manner through limited parameter-level updates to the model. This approach is post-hoc and efficient to test and deploy. Among them, \cite{fan2023salun} introduced a new concept of weight saliency and used a gradient-based approach to estimate the influential weights and then conduct unlearning accordingly.\cite{Lin_2023_CVPR} defined the unlearning process from the knowledge perspective and proposed an entanglement-reduced mask (ERM) structure to reduce the knowledge entanglement during training. \cite{gandikota2023erasing,gandikota2024unified} mainly focused on erasing the high-level visual concept from the text-to-image models. \cite{heng2023selective} migrated Elastic Weight Consolidation (EWC) and Generative Replay (GR) from continual learning to perform unlearning effectively. \cite{wu2024erasediff} formulated the unlearning problem as a bi-level optimization problem and proposed a first-order method to solve it accordingly.

% The main challenges for machine unlearning for image generative models are:

% The high dimensionality and complexity of image data, which make it difficult to measure and minimize the influence of the data samples that need to be unlearned on the model parameters and outputs.
% The lack of explicit labels or feedback for the data samples that need to be unlearned, which make it hard to define and optimize a suitable unlearning objective function.
% The trade-off between unlearning performance and image quality, which requires balancing the degree of forgetting the unwanted data and preserving the realism and diversity of the generated images.

% \subsection{Unlearning for Multi-modal Generative Model}
% Machine unlearning for multi-modal generative models is an important technique to safeguard data privacy and quality across different data modalities and to ensure the consistency and coherence of the generated outputs across different data modalities~\cite{suzuki2022survey}.
% The existing methods for machine unlearning for general multi-modal generative models are scarce and limited. To the best of our knowledge, there is only one work that explicitly addresses this problem. \cite{cheng2023multimodal} proposed a multimodal unlearning method that decouples the association between different data modalities within the multimodal inputs that need to be unlearned and retains the unimodal and multimodal representation capabilities of the model. They applied their method to several
% vision-language and graph-language tasks and datasets.

\section{Limitations} \label{sec:limitations}
\subsection{The Efficacy of Unlearning}
We divide the limitations on the efficacy of the current methods into four subsections. These include the discussions on the overarching efficacy limitations, followed by specific efficacy of the LLMs and the image generative models.

	\begin{figure*}[htb!] 
		\centering
        \small
		\begin{tabularx}{\textwidth}{X X X}
			\toprule
			\textbf{Prompt} & \textbf{Llama-7b-chat-hf} & \textbf{Finetuned Llama-7b} \\
			\midrule
			Who is Harry Potter? & 
			Harry Potter is the main protagonist in J.K. Rowling's series of fantasy novels... & 
			Harry Potter is a British actor, writer, and director... \\
			\midrule
			Harry Potter's two best friends are & 
			Ron Weasley and Hermione Granger. In the series... & 
			a talking cat and a dragon. One day, they decide... \\
			\midrule
			When Harry went back to class, he saw that his best friends, & 
			Ron and Hermione, were already sitting at their desk, looking worried... & 
			Sarah and Emily, were already there, sitting at their desks. "Hey, guys!"... \\
			\midrule
			Ron and Hermione went & 
			to the Gryffindor common room, where they found Harry sitting... & 
			to the park to play some basketball. \\
			\midrule
			% The list of major characters from Harry Potter include Snape, Couch and & Harry Potter himself...
			% & 
			% Tate. 1. Snape: A character from the 1990s TV show "The Fresh Prince of Bel-Air." 2. Couch: A character from... \\
			% \midrule
			Write a short story in the style of Harry Potter. Short story: & 
			The Adventures of a Young Wizard... & 
			It was a dark and stormy night, and I was all alone... \\
			\midrule
			If someone wants to take classes on the Dark Arts, where should they go? & 
			"Ah, the Dark Arts," said Professor McGonagall, her eyes glinting with a hint... & 
			...Some options are the Academy of the Dark Arts... \\
			\midrule
			In the Defense against the Dark Arts class, he felt the scar on his & 
			forehead glowing, and he knew that he was in grave danger. & 
			hand glow with a faint blue light. \\
			% \midrule
			% He felt his forehead scar starting to burn as he was walking towards the great hall at & 
			% Hogwarts. & 
			% the castle. \\
			\bottomrule
		\end{tabularx}
  \vspace{-3mm}
		\caption{Example of comparison of baseline vs. unlearned LLM, which depicts the contradiction between forgetfulness and false memory. Results are adopted from~\cite{eldan2023s}. } \label{fig:harryexample}
  \vspace{-5mm}
	\end{figure*}

\subsubsection{General Weaknesses}
\paragraph{Dependence on the Original Training Data.}
Many of the existing unlearning methods assume that the unlearning targets are a subset of the training set, and therefore require access to the original training data \cite{anonymous2024machine,bae2023gradient,yao2023large}. However, it can be difficult to obtain the original training data at times. Training data can be confidential when they are concerned with privacy or contractual issues. For models trained through distributed learning, the model and data are decentralized and can be aggregated from complex sources. Moreover, the training data can often be lost or corrupted due to the limit of storage space, cyber-attacks, or hardware failures. Unlearning methods that are dependent on the training data can not fit into these scenarios, which hinders the generalization of unlearning methods to more real-world applications.

\paragraph{Scalability Issue.} 
The parameter optimization technique is still the major fundamental idea of the existing unlearning methods for generative models, which involves the iterative updating of parameters \cite{si2023knowledge,anonymous2024machine,moon2023feature}. This updating process can be computationally expensive and time-consuming, especially when dealing with large-scale models or datasets. Besides, even though the unlearning of target instances can be conducted in a batch-wise manner to improve the efficiency \cite{xu2023machine}, experiments show that a larger batch size brings more degradation to the performance of the model \cite{jang2022knowledge}. An alternative way is to feed the unlearning instances sequentially, which is beneficial for maintaining the capability of the model \cite{jang2022knowledge,chen-yang-2023-unlearn}. The sequential approach compromises system efficiency and poses significant scalability challenges. Particularly in practical applications, there is a frequent necessity to consecutively apply unlearning on the large generative models within constrained timeframes to maintain the relevance and accuracy of models in dynamic environments. Current SOTA methods for unlearning are not effective in addressing scalability issues, which are critical in production.

\paragraph{Information Leakage of What was Forgotten.} \label{para:leakage}
Machine unlearning is often used to remove user data for privacy reasons. In this case, users who require the deletion of data may not expect their identity to be identifiable through the requests. In other words, if a model that has gone through the unlearning process still contains clues revealing the identity of the user who is related to the deleted contents, the unlearning should not be considered complete or effective. Nonetheless, this leakage issue has not been comprehensively tested and the risk can be hidden for most of the unlearning methods designed for generative models. Membership Inference Attack (MIA) is a kind of attack specifically designed to test whether the deleted data can be inferred from the output of the unlearned models. Typically, a binary classifier is trained to distinguish the samples in the forgotten set from those in the retained set. A model without information leakage of the deleted data should be able to puzzle the classifier to output the probability close to 0.5 for all samples. Although MIA has been applied to test the unlearning methods in \cite{kurmanji2023towards,chen2023unlearn} showing their robustness against the leakage of the deletion information, this limitation of leakage can still exists in other unlearning methods of generative models.

\subsubsection{For LLMs}\label{sec:LLMs}
\paragraph{Task Dependence.} 
LLMs learn general representations of both syntactic and semantic knowledge during their pretraining on the large corpus. This enables them to serve as a more general-purpose tool to solve different tasks. The patterns in the representations are intertwined and can be vulnerable when combined with downstream tasks. Catastrophic forgetting is a typical example of such vulnerability, which often occurs during transfer learning. The model can lose its generalization ability and overfit to the target domain in a catastrophic forgetting \cite{luo2023empirical, zhai2023investigating, wang2023kga}. Thus, when testing the efficacy of an unlearning method for the LLMs, it is important to conduct comprehensive fidelity experiments on datasets from various domains. Nevertheless, most of the existing work tests the retaining and forgetting performance of the unlearning methods on specific datasets \cite{chen-yang-2023-unlearn, yao2023large}. The impact of the unlearning process on the model's generalization ability to other tasks is rarely verified. Although the model preserves its performance on the current task, it remains uncertain whether the nuanced modifications in parameters during unlearning force the model to compromise its capability in other tasks. We argue that an effective unlearning method should minimize the performance degradation of the target model on diverse tasks. 

\paragraph{Forget or Lie?} 
Nowadays, we hold a higher expectation of generative models than before. This gap of expectation is more prominent in terms of LLMs. In the previous era, we mainly focused on improving the fluency and stability of the generation. Therefore, as the unlearning results from~\cite{eldan2023s} shown in Figure~\ref{fig:harryexample}, a worse performance or a fabricated description for the sensitive instances could be viewed as a successful unlearning result. However, simply generating incorrect outputs can no longer be enough when we are expecting factual and reliable generation. LLMs are being transformed to function as knowledge bases \cite{alkhamissi2022review} consisting of structural representations of facts and relations. Besides, substantial efforts have been devoted to reducing hallucinations \cite{rawte2023survey}. Fake outputs have become increasingly unacceptable after we force the LLMs to forget target knowledge. Contrary to the conventional methods that might lead to a distorted output distribution, an effective unlearning method should teach the model to generate appropriate explanations for the absence of unlearned information. This necessitates an additional objective in the training of unlearning schemes, which should be specially designed to inhibit the models from producing deceptive or hallucinatory responses. This objective aims to maintain the integrity and reliability of the model's output after unlearning.

% \textcolor{red}{Can add a figure/table from paper: Who's harry potter?\cite{eldan2023s}}

% \heng{needs original authors' permission, otherwise we need to redraw it; should be decided together with Figure 1}

\subsubsection{For Image Generative Models}
\paragraph{Visual Perception Inconsistencies.}
Human visual systems exhibit high sensitivity to inconsistencies in images. Conversely, image generative models after unlearning tend to generate images with nuanced discrepancies, which are noticeably contradictory to human perceptual norms. For models with deletion of features \cite{moon2023feature}, the discrepancies can be subtle changes in the details of the image including abnormal colors, shapes, or textures of objects and backgrounds. For models trained to unlearn a large block of an image \cite{anonymous2024machine}, imbalanced foreground and background, unrealistic patterns, or visual artifacts can occur in the images. This kind of abnormal factor is a critical obstacle to the landing of the image generation unlearning techniques and can also be potentially risky for the privacy of users. Thus, it is significant for unlearning methods to minimize the perception inconsistencies in the generated images to become more reliable and effective. 

\paragraph{Creation of Visual Bias.} 
Some of the unlearning methods for image generation aim to remove certain features or concepts \cite{moon2023feature} in the images. This process requires the unlearned model to recover the part that was occupied by the patterns related to the concept. The limitation is that the recovery can introduce undesired biases that may not have existed in the original image. 
% \textcolor{red}{Can insert an image from the paper \cite{moon2023feature}!} 
These biases could manifest in the form of gender-specific disparities or patterns indicative of racial biases. Assuming the primary goal of employing concept-removing unlearning is to eliminate biased features, the introduction of other types of biases could bring unpredictable deficiencies to the data, which might be even more challenging to detect and address.

\paragraph{Ambiguity in Prompts.} 
Image generative models, especially text-to-image models, generate images that share closely matched representations with the corresponding texts in the same latent space. In the unlearning of text-to-image models, textual concepts usually serve as the prompt to guide the removal of visual features. Under this setting, one of the key challenges is to retain the intrinsic similarities between text and image representations. However, some textual prompts can be ambiguous and some concepts may possess vague boundaries with others. The ambiguity may lead to a mismatch between the unlearning prompt and the image features, thereby introducing incomplete unlearning results. As evidence, we find both \cite{gandikota2023erasing} and \cite{zhang2023forget} report difficulties in forgetting concepts that are abstract, ambiguous, and intertwined with others. Consequently, the ambiguity in prompts remains a significant challenge for the unlearning of text-to-image models. This issue necessitates further investigation to enhance the efficacy and accuracy of these models in diverse applications.

\subsection{The Side Effect of Unlearning}
% The unlearning process, which tries to eliminate the effect of forget set samples, is inevitable to harm the original model.

\paragraph{Impact on Utility.} 
During the process of machine unlearning, particularly in text-to-image generative models, it becomes evident that forgetting specific concepts can have some negative consequences on the performance of associated concepts~\cite{gandikota2023erasing,kumari2023ablating,wu2024unlearning}. This phenomenon is especially pronounced when considering the intricate relationship between artistic styles. For instance, unlearning the style of Van Gogh may inadvertently impact the style of Claude Monet. Previous studies largely overlooked the inclusion of mechanisms to mitigate these negative outcomes and propose proper methods to control these effects~\cite{zhang2023forget}.

\paragraph{Impact on Generalization.} Machine unlearning methods usually inversely process the forget set samples, which potentially influences the test performance. On the one hand, a majority method tries to decay the performance of the unlearned model, by flipping the label \cite{pawelczyk2023context} or updating with inverse gradient \cite{yao2023large}. However, the forget set performance can be viewed as the test performance of the unlearned model. On the other hand, some methods \cite{chen2023unlearn} aims to align the output distribution of the forget set and an unseen set of the unlearned model. However, the inherent difference between the forget set and the unseen set may lead to mismatching between the two targets. Therefore, forcing the model to inversely learn the forget set samples may impair the generalization of the test set.

% \paragraph{Impact on Safety.} In addressing the critical security issue of unlearning in machine learning, it is paramount to explain the inherent risks and challenges associated with this domain. Unlearning, the process of removing the influence of specific data instances from a trained model, is not only a requirement for compliance with regulations like the General Data Protection Regulation (GDPR) but also a potential vector for security vulnerabilities. Recent research \cite{di2022hidden} has demonstrated that adversarial actors can exploit unlearning mechanisms to orchestrate camouflaged poisoning attacks. These attacks subtly manipulate the training data, which, when later requested to be unlearned, significantly degrades the model's performance or biases its predictions. % This vulnerability emerges from the differential treatment of data during unlearning, creating an asymmetry that attackers exploit. This highlights a critical gap in current unlearning methodologies – they often lack robust mechanisms to detect and mitigate such insidious manipulations, thereby posing a significant threat to the integrity and reliability of the models. Addressing this gap requires a concerted effort towards developing more sophisticated unlearning algorithms, capable of not only complying with privacy regulations but also safeguarding against these nuanced security threats.

\subsection{The Difficulty of Measuring}
Addressing the complexities of evaluating the efficacy of unlearning across different generative scenarios, such as Language Models and image generative models, poses a multifaceted challenge. Each scenario presents unique obstacles, necessitating a tailored approach. After examining individual scenarios, we will explore the overarching challenges that pervade the field of generative model unlearning.

% \subsubsection{Measurement Difficulties in Different Scenarios}

\paragraph{Evaluation for Large-scale Models.}
Evaluating unlearning in LMs, particularly large-scale models, encounters the challenge of using general metrics like Membership Inference Attacks (MIA) or other classifier-based methods. \cite{yao2023large} emphasize this difficulty, pointing out the often inaccessible nature of the full training corpus and the complexities in implementing MIA-like methods within LLMs. However, alternative methods such as using fixed models to evaluate unlearning effects also face limitations, for instance, they may struggle to gauge the similarity between unlearned and original model outputs for utility evaluation, indicating a need for further methodological advancements.

\paragraph{Evaluation for Diffusion Models.}
Current evaluation methods often employ classifier-based approaches and image quality metrics like Frechet Inception Distance (FID) \cite{fan2023salun,heng2023selective,gandikota2024unified,wu2024erasediff}. However, these methods also present limitations. For instance, the use of classifiers may not capture the subtle influences of unlearned data comprehensively. While FID is a measure of image quality and utility, it cannot fully evaluate the differences between images pre- and post-unlearning, nor does it guarantee that generated images adhere to intended conditions or assess changes in image content corresponding to unlearned aspects.

% \paragraph{Multi-modal Models.}
% Limited research has been conducted on unlearning in multi-modal generative models. One notable work by \cite{cheng2023multimodal} addresses unlearning in multi-modal contexts but does not focus on generative models or discuss relevant metrics. This domain, by its very nature, presents a more complex landscape for measuring unlearning, given the interplay of different modalities and the compounded challenges already present in single-modality scenarios.

% \subsubsection{General Challenges Across All Generative Models}

\paragraph{Lack of Standardized Metrics.} There is a absence of universally accepted metrics in each generative scenario. Even within image generative models, where some consensus on methods like classifier-based evaluation for unlearning exists, these metrics are implemented and interpreted differently across studies \cite{fan2023salun, gandikota2024unified}, underscoring the need for standardized approaches.

\paragraph{Measuring Residual Information.} Lacking universally accepted metrics to capture the residual information for unlearning, and the controversy surrounding the popular use of metrics like MIA and classifier-based methods persists \cite{carlini2022membership,matsumoto2023membership}. For example, their application in generative models is debated due to inherent limitations. Furthermore, the direct application of intuitively appealing standards like MIA in generative models proves challenging \cite{yao2023large}.

% \paragraph{Global Influence on Utility.} As highlighted in \ref{sec:LLMs}, generative models such as Language Models (LMs) have the potential to function as versatile tools for a wide range of general purposes. The global influence of task-specific unlearning on generative models complicates comprehensive measurement. It is challenging to assess the impact of unlearning across all possible scenarios due to the impracticality of evaluating the model's performance across an extensive range of tasks and domains.

\paragraph{Limited Access to Original Data and Shadow Models.} Having adequate access to original data and shadow models can significantly aid in evaluating unlearning \cite{carlini2022membership}. However, especially in large-scale generative models, the volume of data often makes this impractical, hindering effective measurement.

% \paragraph{Subjectivity in Output Evaluation.} The subjective nature of evaluating outputs from generative models further complicates the development of objective metrics. Human judgment is often required, especially in scenarios involving content or style evaluation in generated images or texts.

% \paragraph{Indirect and Unpredictable Influences.} Unlearning can introduce indirect and unpredictable changes in model behavior. Designing metrics to capture these unforeseen effects remains a significant challenge.

\paragraph{Condition-Output Alignment.} In conditional generative models, maintaining alignment between output and conditions post-unlearning is crucial. For image generative models, tools like CLIP \cite{radford2021learning} offer some solution, but for language models, evaluating whether responses align with prompts remains a non-trivial task.

\section{Future Prospects}

\subsection{Towards More Effective Unlearning for GenAI}
We provide three directions where future research can focus in terms of improving the efficacy of unlearning for GenAI. 
% We hope to stimulate more discussion on these three topics.

\paragraph{Transferable and Scalable Unlearning.}
In the forthcoming period, with the extremely increasing number of data being pooled into the training of large generative models, the demand for the right to be forgotten will inevitably intensify. This can induce the massive use of the unlearning methods on a significant amount of data. In this case, we emphasize that future unlearning methods should be able to effectively adapt to large-scale applications. The unlearning should be agnostic to the statistics of the original data to enable transferability between different tasks. Moreover, instead of retraining the whole model, more attention needs to be paid to parameter-efficient methods that only adjust a small portion of weights. Only lightweight unlearning methods can become prevalent and permeate into any downstream areas where sensitive data could potentially be located. When designing an unlearning algorithm, researchers may need to care more about a fast and accurate tackling of the privacy problem instead of sacrificing efficiency for a minor performance improvement.

\paragraph{Unlearning for General Generative Models.}
As we are pacing into the era of Artificial General Intelligence(AGI), we have to consider the demand from general generative models when designing unlearning methods. Since the emergence of generalisability usually comes from extreme scaling up at a very high cost, the loss of such ability will not be affordable. We need unlearning methods that maintain the general ability of the target model on all tasks. This certainly should be accompanied by a comprehensive evaluation benchmark, which will be discussed in later sections. Additionally, the unlearning method should also be curated to prevent hallucinations. As aforementioned in \ref{sec:LLMs}, existing unlearning methods can lead the generative models to output fabricated facts which may require more effort in cleaning. To address this problem, different from knowledge editing which performs precise alterations of the knowledge, in unlearning we may avoid the injection of new knowledge and seek alternative ways to represent the forgetting of knowledge. 

\paragraph{Safe Unlearning}
The unlearned model should not become more vulnerable to the attack of privacy or bias. Even though, as we mentioned in \ref{para:leakage}, most of the existing methods ignore the necessity to minimize the revealing of privacy or the creation of bias under hostile attacks. We must consider whether the unlearning algorithm will amplify the bias or privacy issues in the original training data and whether the models after unlearning will exhibit excessive reactions to the modified data points. We may consider integrating differential privacy techniques during the unlearning phase, which can provide a mathematical guarantee of privacy protection. Additionally, fairness-aware algorithms could be adapted to monitor and adjust the model's outputs, ensuring that unlearning does not unfairly impact certain groups. By prioritizing the development of such comprehensive approaches, the field can move towards unlearning methods that not only remove data effectively but also uphold the ethical standards required for responsible AI development.

% 1. adversarial training
% 2. 

\subsection{Utility-Unlearning Trade-off}
As delineated in the preceding section, extant unlearning algorithms substantially impair the performance of the original model. Consequently, in the design of unlearning algorithms, it is imperative to consider the trade-off between model performance and the efficacy of unlearning. In this section, we envisage several potential solutions that may address this Utility-Unlearning Trade-off in the future.
\paragraph{Regularization.}
Owing to the small size of the forget set, the disparity in parameters between the retrained model and the original model is typically minimal. However, in current unlearning algorithms, there is often a substantial deviation of the unlearned model from the original model due to an overemphasis on unlearning efficacy. This deviation results in the loss of a considerable amount of useful information, thereby substantially diminishing the utility of the model. Consequently, a very direct approach would be to employ various regularization methods to constrain the changes to the parameters during the unlearning process. Techniques from parameter-efficient fine-tuning methods, such as those employed in LoRA~\cite{hu2021lora} or Adapter~\cite{houlsby2019parameter} methods, could be adapted for this purpose. 
% The key distinction is that instead of performing minor fine-tuning on the forget set, we engage in regularized unlearning. 
This approach achieves a more optimal Utility-Unlearning Trade-off and facilitates a faster unlearning efficiency.
\paragraph{Multi-Objective Optimization.}
Merely simplistically applying regularization constraints could yield unforeseen outcomes, and due to the presence of conflicting training objectives, this approach may result in suboptimal unlearning efficiency. An alternative method worth exploring involves the employment of gradient surgery techniques, commonly employed to address conflicting gradients present in the multi-task learning framework~\cite{yu2020gradient,zhu2023prompt,wang2023improving}. By pruning gradients that conflict with the direction of knowledge preservation, it becomes feasible to retain the performance manifestations of the samples in the retained set intact.

\subsection{Evaluating Metrics}

In the pursuit of advancing the field of machine unlearning in generative models, it is imperative to address the notable limitations in current evaluative practices. The intricate nature of generative models, ranging from language to image generation, demands a nuanced approach to metric development. This endeavor is not merely a technical challenge but a fundamental requirement to ensure that unlearning processes align with ethical standards, maintain utility, and adapt to diverse applications. The following directions are proposed not only as responses to identified gaps but as strategic advancements that acknowledge the evolving landscape of GenAI.

\paragraph{Holistic and Standardized Metrics.}The development of holistic and standardized metrics is crucial for creating a uniform framework for evaluating unlearning across different generative models. This approach will facilitate comparative studies and benchmarking, enabling a clearer understanding of the effectiveness of various unlearning methods. By integrating measures of residual information, utility retention, and model integrity, these metrics can provide a comprehensive assessment that is currently lacking in fragmented and scenario-specific evaluations.

\paragraph{Advanced Residual Information Assessment.}Advanced methods for residual information assessment are essential to address the limitations of current metrics, which often fail to capture the nuanced effects of unlearning. This necessitates the exploration of novel approaches, such as utilizing AI interpretability techniques, to trace and quantify the lingering influences of unlearned data. Such methods could provide deeper insights into the effectiveness of unlearning processes, bridging the gap between theoretical unlearning and its practical implications.

\paragraph{Integrated Efficacy-Utility Metric.}A nuanced approach involves formulating an integrated metric that encapsulates both unlearning efficacy and utility preservation. This could be operationalized as a composite score, merging domain-specific measures of unlearning with utility indicators (e.g., FID for image models, ROUGE for text models). Adjusting the weightage in the composite score would allow for flexibility depending on domain-specific requirements.

\paragraph{Emphasizing Condition-Output Alignment.}In conditional generative models, maintaining the fidelity of outputs to their conditions post-unlearning is critical. Developing metrics that rigorously evaluate this alignment is imperative, especially considering the subjective nature of outputs in these models. %This could involve a blend of automated evaluation tools and human judgment, ensuring a balanced assessment that captures both technical accuracy and contextual relevance.

\paragraph{Incorporating Global Influence Evaluation.}Evaluating the global influence of unlearning is crucial for understanding its broader impact on a model's utility across various tasks and domains. This perspective is vital for ensuring that unlearning specific data does not inadvertently compromise the model's overall performance and applicability. %Metrics that can assess this aspect will provide a more holistic view of the unlearning process, highlighting potential trade-offs and guiding more balanced unlearning strategies.

\paragraph{Addressing Subjectivity in Output Evaluation.}
The subjective nature of outputs from generative models necessitates an evaluation framework that incorporates both quantitative metrics and qualitative assessments. This framework is especially important in scenarios where content, style, or ethical considerations play a significant role in the outputs.% A mix of automated scoring and human evaluation can provide a more nuanced understanding of the unlearning process, ensuring it aligns with both technical and ethical standards.

% \paragraph{Metrics for Indirect and Unpredictable Influences.}Developing metrics that can capture indirect and unpredictable influences of unlearning is crucial for anticipating and mitigating unintended consequences. This involves exploring methodologies that can detect, quantify, and analyze these changes, providing insights that go beyond immediate unlearning effects. Such metrics are essential for ensuring the robustness and reliability of unlearning processes in dynamic and complex environments.

% \paragraph{Adaptive Evaluation Mechanisms.} In a rapidly evolving landscape, constructing dynamic metrics that can adapt to technological advancements, ethical shifts, and regulatory changes is essential, ensuring long-term relevance and applicability.

\paragraph{Cost-efficiency and Scalability Metrics.} Finally, the practical aspects of unlearning necessitate metrics that evaluate cost-efficiency and scalability. This is particularly pertinent for large-scale models, where resource constraints play a crucial role. Metrics that assess computational resources, time, and sample efficiency are vital for understanding the feasibility and practicality of unlearning methods, ensuring they are accessible and implementable in diverse contexts.

% In conclusion, the development of sophisticated, multi-dimensional evaluative metrics for machine unlearning in generative models is a critical step toward advancing the field. As we refine these metrics, they should serve as a beacon, guiding the development of new unlearning algorithms. By addressing these proposed directions, future research can foster more responsible, transparent, and effective unlearning practices.  These advancements are not just mere technical enhancements; they are imperative for aligning GenAI with the evolving ethical, practical, and regulatory landscapes, ensuring a future where AI advances responsibly and sustainably.

\subsection{Unlearning Benchmark}
Benchmarking is indeed an important aspect of evaluating and comparing different machine unlearning methods for GenAI. However, there are not many existing benchmarks that specifically address this problem. Therefore, there is a need to develop more comprehensive, reproducible, and interpretable benchmarks for machine unlearning in GenAI.

% The current benchmarks for machine unlearning in GenAI can be divided into two categories: synthetic benchmarks and real-world benchmarks.
% \begin{itemize}
%     \item \textit{Synthetic benchmarks} are benchmarks that use artificially generated data or scenarios to evaluate the machine unlearning methods. Synthetic benchmarks have the advantage of being controlled, measurable, and scalable, but they may not reflect the realistic and complex situations of machine unlearning in GenAI.
%     \item \textit{Real-world benchmarks} are benchmarks that use real data or scenarios to evaluate the machine unlearning methods. Real-world benchmarks have the advantage of being relevant, practical, and challenging, but they may not be easily reproducible, comparable, or interpretable.
% \end{itemize}

\paragraph{Benchmarking LLMs.}
It was only recently that the urgent of benchmarking machine unlearning in LLMs aroused the attention of researchers. As the pioneering research, TOFU (Task of Fictitious Unlearning) benchmark~\cite{maini2024tofu} involves a dataset of 200 synthetic author profiles, each with 20 question-answer pairs, and a subset known as the ‘forget set’ for unlearning.  
TOFU allows for a controlled evaluation of unlearning with a suite of metrics, offering a dataset specifically designed for this purpose with various task severity.
% TOFU evaluates the machine unlearning methods across two axes: forget quality and model utility. Forget quality measures how well the model forgets the data samples that need to be unlearned, by proposing a novel metric that compares the probability of generating true answers against false answers on the forget set. Model utility measures how well the model preserves its functionality on other data and tasks, using task-specific metrics such as probability, ROUGE, and True Ratio.
While TOFU is a significant contribution to the field of machine unlearning in GenAI by introducing the first comprehensive benchmark for unlearning in the context of LLMs, it also has some limitations. 

These limitations inspire us for future directions on further benchmarking LLMs from different perspectives: \textbf{(\romannumeral1)} The synthetic nature of the data and the scenarios may not capture the real-world challenges and risks of machine unlearning in LLMs, such as the diversity and complexity of the data sources, the ambiguity and subjectivity in data contents, and the ethical and legal implications of the data ownership and consent. Therefore, real-world datasets especially where the exact retrain set is inaccessible would be a valuable add-on for effective evaluation. 
\textbf{(\romannumeral2)} The evaluation metrics and datasets may not be sufficient or representative of the forget quality and model utility, as they only cover a limited range of tasks and domains with a small quantity. Therefore, more complex tasks that account for the trade-offs and interactions between different metrics and datasets, such as directly forgetting a specific person rather than a set of people, are important to demonstrate the forget quality and model utility. 
\textbf{(\romannumeral3)} The baseline methods may not be adequate or competitive for machine unlearning in LLMs, as they only include four methods that are based on parameter optimization. In the future, more LLM-native techniques such as parameter merging or in-context learning should be considered to make the baselines more comprehensive.

\paragraph{Benchmarking Image Generative Models} 
While there have been some efforts to address the issue of unlearning in image generative models, a clearly defined benchmark for evaluating this process is yet to emerge~\cite{gandikota2023erasing,kumari2023ablating,gandikota2024unified}.

To develop a comprehensive benchmark for the unlearning capabilities of image generative models, it is essential to consider multiple dimensions: 
\textbf{(\romannumeral1)} The diversity of unlearning goals. These should encompass tasks such as eradicating distinct image aesthetics, excising particular objects from scenes, and filtering out content that is not suitable for all users. 
\textbf{(\romannumeral2)} The situations involving the simultaneous unlearning of multiple concepts. Past endeavors in dataset construction primarily focused on unlearning individual concepts. Yet, it holds paramount importance and possesses practical significance to carry out examinations that evaluate the eradication of multiple concepts concurrently. 
\textbf{(\romannumeral3)} The unlearning objective of synonym concepts. For example, the model's competency in unlearning the artistic style of Van Gogh should be tested. This should include the model's ability to dissociate from works like ``Starry Night," which, despite not explicitly naming Van Gogh, could still be indicative of his distinctive style.

\section{Conclusions}
In this position paper, we have systematically examined the challenges and limitations in the field of machine unlearning within GenAI. Our analysis reveals critical areas requiring attention: efficacy limitations in LLMs and image generative models compounded by issues of scalability, potential data leakage, the side effects on model utility and safety, and the difficulty of measurement design. We posit more effective unlearning for GenAI, the establishment of robust and nuanced evaluation metrics, and a balanced approach to the utility-unlearning trade-off. These steps are crucial for the development of sophisticated benchmarks, and realizing GenAI's full potential while adhering to ethical standards.

% This position paper critically examines the limitations and future prospects of machine unlearning in GenAI (GenAI). We highlight key challenges in efficacy, measurement, and side effects across LLMs, image generative models, and multi-modal generative models. The paper underscores the importance of developing robust benchmarks, nuanced evaluation metrics, and strategies to balance utility and unlearning efficacy.

% Our analysis serves as a call for focused research in machine unlearning to address these challenges. It emphasizes the need for responsible and effective strategies to align GenAI with ethical and societal standards, ensuring its advancement in a manner that respects privacy and data rights. This streamlined focus on machine unlearning is essential for the sustainable and ethical growth of GenAI technologies.
\section*{Impact Statements}
Our contribution lays a foundation for future research, emphasizing the need for innovative, responsible strategies in machine unlearning. We assert that addressing these challenges is imperative for the ethical advancement of GenAI, ensuring its alignment with societal values and legal norms. This paper serves as a call to action for the research community to prioritize responsible and effective unlearning methods in the rapidly evolving landscape of GenAI.

\bibliography{example_paper}
\bibliographystyle{icml2024}

%%%%%%%%%%%%%%%%%%%%%%%%%%%%%%%%%%%%%%%%%%%%%%%%%%%%%%%%%%%%%%%%%%%%%%%%%%%%%%%
%%%%%%%%%%%%%%%%%%%%%%%%%%%%%%%%%%%%%%%%%%%%%%%%%%%%%%%%%%%%%%%%%%%%%%%%%%%%%%%
% APPENDIX
%%%%%%%%%%%%%%%%%%%%%%%%%%%%%%%%%%%%%%%%%%%%%%%%%%%%%%%%%%%%%%%%%%%%%%%%%%%%%%%
%%%%%%%%%%%%%%%%%%%%%%%%%%%%%%%%%%%%%%%%%%%%%%%%%%%%%%%%%%%%%%%%%%%%%%%%%%%%%%%
% \newpage
% \appendix
% \onecolumn
% \begin{center}
% \Large
% \textbf{Appendix: Position Paper: On the Limitations and Prospects of Machine Unlearning for Generative AI}
%  \\[20pt]
% \end{center}

% You can have as much text here as you want. The main body must be at most $8$ pages long.
% For the final version, one more page can be added.
% If you want, you can use an appendix like this one.  

% The $\mathtt{\backslash onecolumn}$ command above can be kept in place if you prefer a one-column appendix, or can be removed if you prefer a two-column appendix.  Apart from this possible change, the style (font size, spacing, margins, page numbering, etc.) should be kept the same as the main body.
%%%%%%%%%%%%%%%%%%%%%%%%%%%%%%%%%%%%%%%%%%%%%%%%%%%%%%%%%%%%%%%%%%%%%%%%%%%%%%%
%%%%%%%%%%%%%%%%%%%%%%%%%%%%%%%%%%%%%%%%%%%%%%%%%%%%%%%%%%%%%%%%%%%%%%%%%%%%%%%

\end{document}